\documentclass{article}
\usepackage{ijcai23}

\usepackage{times}
\usepackage{soul}
\usepackage{url}
\usepackage[hidelinks]{hyperref}
\usepackage[utf8]{inputenc}
\usepackage[small]{caption}
\usepackage{graphicx}
\usepackage{amsmath}
\usepackage{amsthm}
\usepackage{booktabs}
\usepackage{algorithm}
\usepackage{algorithmic}
\usepackage[switch]{lineno}

\usepackage{inconsolata}
\usepackage{helvet}  
\usepackage{courier}  
\usepackage{url}  
\usepackage{makecell}
\usepackage{graphicx}  
\usepackage{amsfonts,amssymb}
\usepackage{enumerate}
\usepackage{caption}

\title{Fine-tuned vs. Prompt-tuned Supervised Representations: \\
Which Better Account for Brain Language Representations?}

\author{
Jingyuan Sun
and
Marie-Francine Moens
\affiliations
Department of Computer Science, KU Leuven, Belgium\\
\emails
\{jingyuan.sun, sien.moens\}@kuleuven.be
}

\begin{document}

\maketitle

\begin{abstract}
To decipher the algorithm underlying the human brain's language representation, previous work probed brain responses to language input with pre-trained artificial neural network (ANN) models fine-tuned on NLU tasks.  However, full fine-tuning generally updates the entire parametric space and distorts pre-trained features, cognitively inconsistent with the brain's robust multi-task learning ability. Prompt-tuning, in contrast, protects pre-trained weights and learns task-specific embeddings to fit a task. Could prompt-tuning generate representations that better account for the brain's language representations than fine-tuning? If so, what kind of NLU task leads a pre-trained model to better decode the information represented in the human brain? We investigate these questions by comparing prompt-tuned and fine-tuned representations in neural decoding, that is predicting the linguistic stimulus from the brain activities evoked by the stimulus. We find that on none of the 10 NLU tasks, full fine-tuning significantly outperforms prompt-tuning in neural decoding, implicating that a more brain-consistent tuning method yields representations that better correlate with brain data. Moreover, we identify that tasks dealing with fine-grained concept meaning yield representations that better decode brain activation patterns than other tasks, especially the syntactic chunking task. This indicates that our brain encodes more fine-grained concept information than shallow syntactic information when representing languages.

\end{abstract}

\section{Introduction}
To decipher the mechanism underlying human language representation by analyzing brain activation patterns is a fundamental goal of language neurobiology\cite{schrimpf2021neural,huth2016natural}.
The recent upsurge of ANNs for language processing, exemplified by the Transformer \cite{Vaswani2017Attention,devlin2018bert} and its mutants \cite{ouyang2022training,kojima2022large}, has yielded significant advances in solving a wide range of downstream NLP tasks. 
It is logical to investigate the potential of ANNs to assist in reverse-engineering the algorithms that underlie language representation in the human brain.

Existing studies have employed representations produced by ANN models to explain the brain activities related to language processing \cite{anderson2016predicting,sun2019towards}. 
Some of the best-performing models such as GPT-2 \cite{radford2019language} are found to account for more than 99\% of explainable variance in neural responses for a corresponding linguistic stimulus \cite{schrimpf2021neural}.
Transformer-based large-scale language models show a consistent advantage over other models for predicting and deciphering brain activities \cite{sun2020neural}. 
Among these works, unsupervised methods optimized for language modeling or similar context-predicting goals take a major part in the evaluated ANN models, while supervised models are far less explored. 
Could additional information learned from task supervision lead to better candidate models to account for the brain's language representation?

To study this question, previous work mostly exclusively fine-tuned pre-trained models, producing supervised representations to match with brain activation patterns \cite{gauthier2019linking,oota-etal-2022-neural}. To fit a new NLU task, the full fine-tuning generally updates the entire parameter space of the model which has been proven to distort the pre-traind features \cite{kumar2022finetuning}. 
However, this is potentially inconsistent with the human brain's mechanism which does not require a reformation of the entire brain's language network to learn a single new task.

In this study, we propose to link the human brain and task-supervised artificial representations with prompt-tuning. We take 10 NLU tasks involving diverse linguistic-cognitive skills from widely used NLU benchmarks such as GLUE \cite{wang-etal-2018-glue} and SuperGLUE \cite{sarlin2020superglue}. A pre-trained BERT model is tuned on the 10 tasks with full fine-tuning and prompt-tuning, respectively. The supervised representations generated by the tuned models are then employed to decipher the brain activation patterns through neural decoding. We comprehensively compare the prompt-tuned against the fine-tuned representations. We also demonstrate which task leads to representations that are better at decoding the brain responses. We find that:

\begin{enumerate}[(i)]
\item On the regions of interest (ROIs) comprising the brain language and semantic network\footnote[1]{A brain functional network is a collection of brain regions that consistently show coordinated activities for certain cognitive functions or during behavioral tasks.}, prompt-tuned representations consistently and significantly outperform the fully fine-tuned peers in decoding accuracy.
\item Tuning on tasks dealing with fine-grained concept meaning including Word Sense Disambiguation and Co-reference Resolution yields representations that are better at neural decoding than tuning on other tasks with both fine-tuning and prompt-tuning.

\end{enumerate}

The proposed prompt-tuning-based framework helps a better understanding of the relationship between supervised artificial and brain language representations: the more similar to the brain working mechanism, the better correlated with the brain data. Furthermore, our experimental findings contribute to deciphering the mechanism supporting the human brain's language processing.

\section{Related Work}
\subsection{Fine-tuning and Prompt-tuning}

In recent years, there have been significant developments in pre-trained language models \cite{NEURIPS2020_1457c0d6,sun-etal-2020-distill}, resulting in remarkable enhancements in performance across various NLP tasks. One widely used technique, known as full fine-tuning, involves updating all model parameters for a specific task, as depicted in Figure 1[a]. While full fine-tuning generally yields good task performance, it can be computationally intensive  due to the requirement of storing gradients and optimizer states for all parameters. Moreover, the cognitive plausibility of fine-tuning can be hindered by its potential distortion of general domain knowledge acquired during pre-training.
To address these concerns, an approach called discrete prompting has been developed as an alternative of fine-tuning \cite{10.1145/3560815}. With discrete prompting, the parameters of a pre-trained model are fixed, and an NLP task is solved by querying the model with a discrete natural language prompt. For example, in machine translation, one might prepend a sample like "A lovely day" with a prompt such as "Translate it to Chinese" and ask the pre-trained language model to predict the subsequent tokens.

While discrete prompting does not require additional training and only necessitates storing a single set of model parameters, its task performance can be highly dependent on the design of the prompt template. As a result, it may be sub-optimal compared to fine-tuning in certain cases \cite{qiu2020pre,lester-etal-2021-power}. Due to this limitation, we do not adopt discrete prompting as a baseline in this paper.
Prompt-tuning, an extension of discrete prompting, has been introduced as a technique that fixes the parameters of a pre-trained model while learning continuous prompt embeddings \cite{qin-eisner-2021-learning}. An example of prompt-tuning is prefix-tuning \cite{li-liang-2021-prefix}, which involves adding trainable embeddings called \textit{prefixes} to the original input word embeddings and optimizing them for downstream tasks. Prompt-tuning has shown superior performance compared to discrete prompting on multiple tasks \cite{chen2022knowprompt} and achieves performance levels close to fine-tuning. In this paper, we leverage the characteristics of prompt-tuning to preserve the knowledge learned by a pre-trained model. This enables us to investigate whether the additional supervised information obtained from downstream tasks improves neural decoding performance or not. Specifically, we employ the P-tuning V2 method, as illustrated in Figure 1[b] \cite{liu-etal-2022-p}, which is an optimized version of prefix-tuning claiming comparable performance to fine-tuning across a range of natural language understanding tasks. 

\begin{figure*}[ht]
\centering
\small
\includegraphics[width=5.7in]{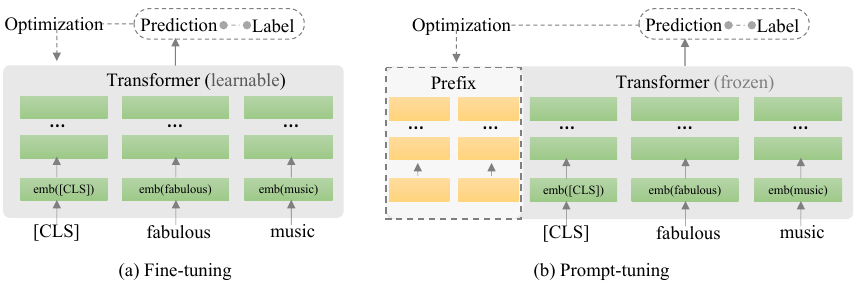}
\caption{Tuning Transformer on downstream tasks [a] with fine-tuning [b] with prompt-tuning.}
\label{fig1}
\end{figure*}


\subsection{Matching Artificial and Neural Representations of Language}

The use of artificial neural network (ANN) representations has proven successful in encoding and decoding neural responses to linguistic stimuli \cite{pereira2018toward,caucheteux-etal-2021-model-based}. Early studies have demonstrated their ability to recover simple verbal stimuli, such as words \cite{pereira2011generating,handjaras2016concepts}, as well as phrases \cite{fyshe2015corpora,huth2016natural}, from brain activation patterns recorded by functional magnetic resonance imaging (fMRI). More recent work have expanded such capability to decoding sentences using similar methods \cite{matsuo2016generating,oota2018fmri,pereira2018toward,wang2022fmri}.

The majority of ANN models evaluated in previous studies have focused on unsupervised representations optimized for language modeling or context-predicting tasks, while supervised models have received comparatively less attention. One relevant research \cite{oota-etal-2022-neural} fine-tuned a pre-trained BERT model on a series of downstream tasks and compared their neural encoding performances. However, since fine-tuning differs from how the human brain adapts to new tasks, we question whether fine-tuning is the most suitable framework for obtaining supervised representations to probe the human brain. To investigate this, we conduct a comprehensive comparison of experiments using fine-tuned and prompt-tuned representations for neural decoding.
It's important to note that our focus in this paper is more on deciphering the information represented in the human brain for language understanding, so we do not conduct neural encoding experiments as done in \cite{oota-etal-2022-neural}.


\section{Methods}

In the following subsections, we will first introduce how prompt-tuning with the P-tuning method works. We then demonstrate how the representations produced by prompt-tuned models are used to build a neural decoder. 

\subsection{Prompt-Tuning}
We build upon the P-tuning method \cite{liu-etal-2022-p} to prompt-tune a pre-trained model on downstream tasks. In Figure \ref{fig1}, we illustrate the mechanism of P-tuning with a Transformer-based language model. P-tuning prepends a continuous trainable prefix to different layers of an ANN to guide its fitting on downstream tasks.

We now explain the mechanism of P-tuning, with tuning a Transformer-based language model on a conditional generation task as an example, following \cite{li-liang-2021-prefix}.
For a given conditional generation task, where the input context is denoted as $x$ and the output token sequence is denoted as $y$, we consider a language model $LM = p_{\phi}(y|x)$, which is parameterized by $\phi$. To facilitate the P-tuning process, we define the variable $z$ as the concatenation of $x$ and $y$, and we represent the corresponding sequence of indices for $x$ and $y$ as $id_{x}$ and $id_{y}$, respectively. At each time step $i$, the activation is denoted as $h_i$, which is obtained by concatenating the activation layers at that specific time step, represented as $h_i=[h_i^1, h_i^2, ... h_i^n]$. Here, $h_i^j$ refers to the activation of the $j$-th Transformer layer at time step $i$. The computation of $h_i$ relies on $z_i$ and the past activations within its context, and it can be formulated as follows: $h_i = \mathbf{LM}_{\phi}(z_i, h)$.
The distribution for the next token is computed as follows: $p_\phi\left(z_{i+1} \mid h_{\leq i}\right)=\operatorname{softmax}\left(W_\phi h_i^{(n)}\right)$. In this equation, $h_i^{(n)}$ represents the hidden states of the last layer within $h_i$. The matrix $W_\phi$ is pre-trained to map these hidden states $h_i^{(n)}$ to logits across the vocabulary.

During P-tuning, a prefix, denoted as $Pr$, is prepended to the input sequence, resulting in the concatenation $z = [Pr; x; y]$. $Pr$ is a continuous trainable embedding, its corresponding sequence of indices is denoted as $Pr_{id_{x}}$ with length $|Pr_{id_{x}}|$. A trainable matrix $M_\theta^{Pr}$, with dimensions $|Pr_{id_{x}}| \times dim(h_i)$, is initialized to store the prefix parameters. And
\begin{equation}
h_i=\left\{\begin{array}{ll}
M_\theta^{Pr}[i,:], & \text { if } i \in {Pr}_{\mathrm{id_x}} \\
\mathbf{LM}_\phi\left(z_i, h_{<i}\right), & \text { otherwise. }
\end{array}\right.
\end{equation}
Training P-tuning is then to optimize the following log-likelihood objective: 
\begin{equation}
\max _\phi \log p_\phi(y \mid x)=\sum_{i \in id_{\mathrm{y}}} \log p_\phi\left(z_i \mid h_{<i}\right)
\end{equation}
with $\phi$ being fixed and $\theta$ being trainable.


\subsection{Neural Decoder}

The neural decoder aims to categorize or reconstruct the perceived stimulus by mining the corresponding neural activities. As shown in Figure 2, we tune the pre-trained Transformer model respectively with fine-tuning and prompt-tuning. The tuned models generate sentence embeddings to represent the sentences in the brain imaging dataset. We then build neural decoders with sentence embeddings to decipher the brain images. Here we use regression-based decoding where a semantic vector is directly estimated from the voxels by regression. 

Specifically, given the voxel matrix $V_d\in \mathbb{R}^{N_E \times N_V}$ and sentence embedding matrix $Z_d\in \mathbb{R}^{N_E \times N_D}$ in the training set, where \(N_E\) represents the number of examples, \(N_V\) represents the number of voxels, and \(N_D\) represents the number of dimensions of the sentence embedding, the regression coefficients of the decoder $W_d$ are estimated by minimizing
\begin{align}
\begin{split}
||W_dV-z_i||_2^2+\lambda ||W_d||_2^2
\end{split}
\end{align}
for each column \(z_i\) in \(Z\), i.e., each dimension of the sentence vectors.
\(\lambda\) is the regularization parameter.

The decoder is trained using a 5-fold cross-validation approach. The regression parameters for each dimension are learned by mapping the brain images of the training sentences to distributed representations. Then, the performance is tested by predicting semantic vectors from the brain images of the left-out testing sentences in each fold. This cross-validation procedure is applied to data from all subjects with each supervised representation. Upon completion of the 5 folds, this process yields decoded semantic vectors for each subject and supervised representation combination.

\begin{figure*}[ht]
\centering
\small
\includegraphics[width=6.6in]{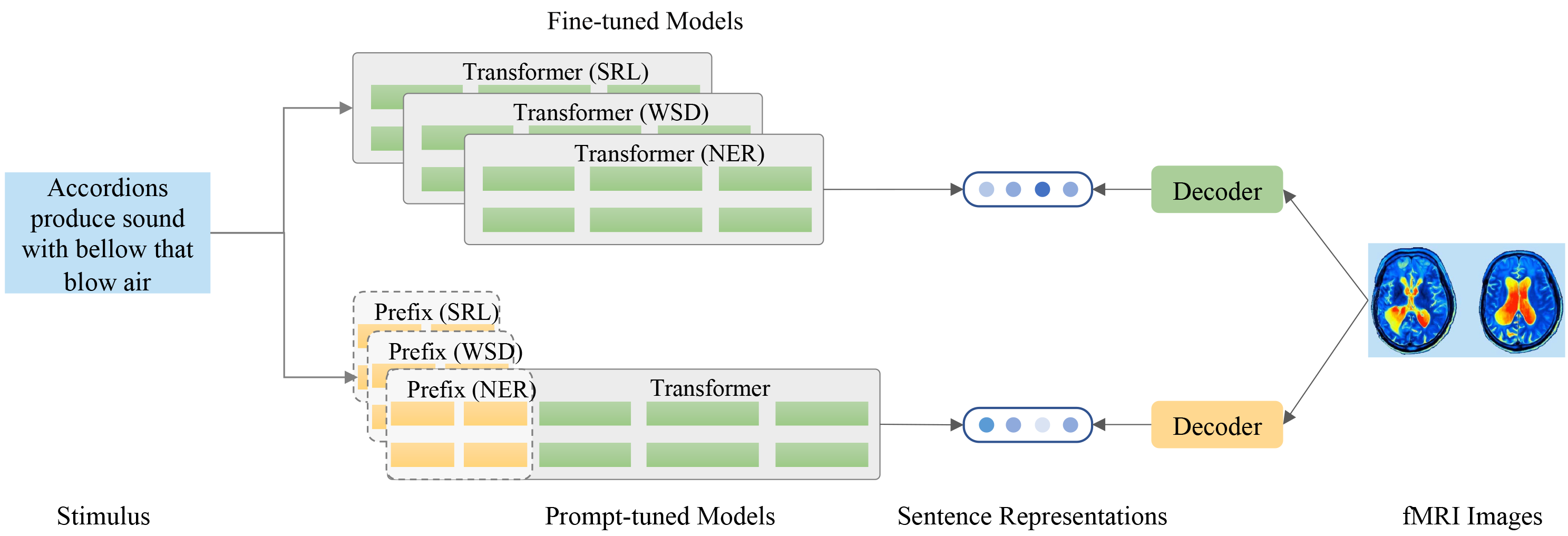}
\caption{Neural decoding with fully fine-tuned and prompt-tuned supervised representations.}
\label{fig2}
\end{figure*}

\section{Experimental Setup}

In this section, we will first briefly introduce the brain imaging dataset we use. We then demonstrate the principle of task selection and details of the selected tasks, as well as tuning on these tasks to generate supervised representations. Codes, checkpoints and other supplementary materials are available online\footnote[2]{\url{https://github.com/soinx0629/sup_fmri_dec_eng}}.

\subsection{Brain Imaging Data}

We utilize the fMRI data published in the study conducted by \cite{pereira2018toward}. 
The fMRI neural activation data were acquired using a 3-Tesla Siemens Trio scanner equipped with a 32-channel head coil. 
Two experiments were conducted in this research, and 5 subjects participated in both of the 2 experiments. The stimuli used in both experiments followed a hierarchical structure of topic-passage-sentence.
In the first experiment, a total of 96 passages, resembling Wikipedia-style content, were presented. Each passage consisted of 4 sentences that focused on a specific concept. These concepts were selected from a pool of 24 broad topics, providing a diverse range of content. The second experiment involved 72 passages, including 48 Wikipedia-style passages and 24 passages written in first/third-person narratives. Each passage contains 3 or 4 sentences about a particular concept unrelated to those in the first experiment. The two experiments were designed to maintain similar semantic relatedness within and between passages/topics.
In this study, we combine the data from the two experiments to make full use of the brain images. There will be imaging data from 5 subjects scanned with 672 sentences presented as stimuli. Such a setting is consistent with previous work \cite{sun2019towards,oota-etal-2022-neural}. The full details about the presented materials and experimental setup can be found online\footnote[3]{\url{https://osf.io/crwz7/wiki/home/}}.

\subsection{Neural Decoding Evaluation}

The evaluation of the decoding results primarily relies on the pairwise matching task, a widely used metric in neural decoding research. The correlation between the decoded vectors and the ground-truth sentence embeddings is first computed. A successful matching is scored when the decoded semantic vectors have a higher degree of similarity with their corresponding brain activation patterns compared to alternative pairings.
Formally, for each possible pair of sentence stimuli \(S_i\) and \(S_j\), let \(X_{S_i}\) and \(X_{S_j}\) denote the brain images of neural responses to \(S_i\) and \(S_j\). Let \(Z_{S_i}\) and \(Z_{S_j}\) denote the sentence embeddings produced by a tuned model,  while  \(D_{S_i}\) and \(D_{S_j}\) denote the decoded semantic vectors from brain images \(X_{S_i}\) and \(X_{S_j}\).
We score 1 for a matching if 
\begin{align}
\begin{split}
corr(D_{S_i},Z_{S_i})+corr(D_{S_j},Z_{S_j})>\\
corr(D_{S_i},Z_{S_j})+corr(D_{S_j},Z_{S_i}),
\end{split}
\end{align}
else 0. \textit{corr} denotes the correlation function employed which is Pearson's correlation in this study. An examination of all possible pairs of sentences will be conducted. The final matching accuracy for each participant is determined by calculating the proportion of correctly matched pairs. The expected chance-level accuracy, assuming a random selection by the model, is 0.50. 

In the experimental results, we will mainly display the pairwise matching performance which is most widely adopted in related work \cite{pereira2018toward,sun2019towards,oota-etal-2022-neural}. Except for pair-wise matching, we also use the MSE loss between the decoding value and ground truth as a supplementary metric. 
Please refer to the appendix for MSE results.

\subsection{Tasks for Tuning}

The tasks chosen for this study are based on the two major principles taking into account the setting of previous work \cite{oota-etal-2022-neural}: 
\begin{enumerate}[(i)]
\item tasks involving a diverse range of cognitive-linguistic skills are desired; 
\item tasks with corpora that are commonly used in popular natural language processing (NLP) benchmarks. 
\end{enumerate}
\begin{figure}[ht]
\centering
\small
\includegraphics[width=2.78in]{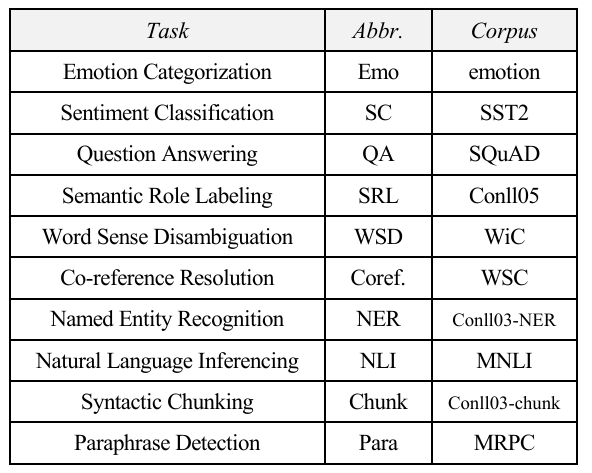}
\caption*{Table 1: 10 tasks used for tuning supervised representations. Abbr. denotes the abbreviation for the task name. 
}
\label{tab2}
\end{figure}

Based on the above principles, we select 
tasks analyzing emotion and sentiment polarity (\textit{Emotion Categorization and Sentiment Classification}), tasks dealing with concept meaning (\textit{Word Sense Disambiguation, Co-reference Resolution and Named Entity Recognition}), tasks that require reasoning semantic relation (\textit{Paraphrase Detection and Semantic Role Labeling}), tasks requiring knowledge and logical representations \textit{(Question Answering and Natural Language Inferencing}) as well as tasks that parse the syntactic structure (\textit{Syntactic Chunking}).
We present these tasks in Table 1.

\subsection{Generating Sentence Embeddings}

\begin{figure*}[ht]
\centering
\small
\includegraphics[width=6.9in]{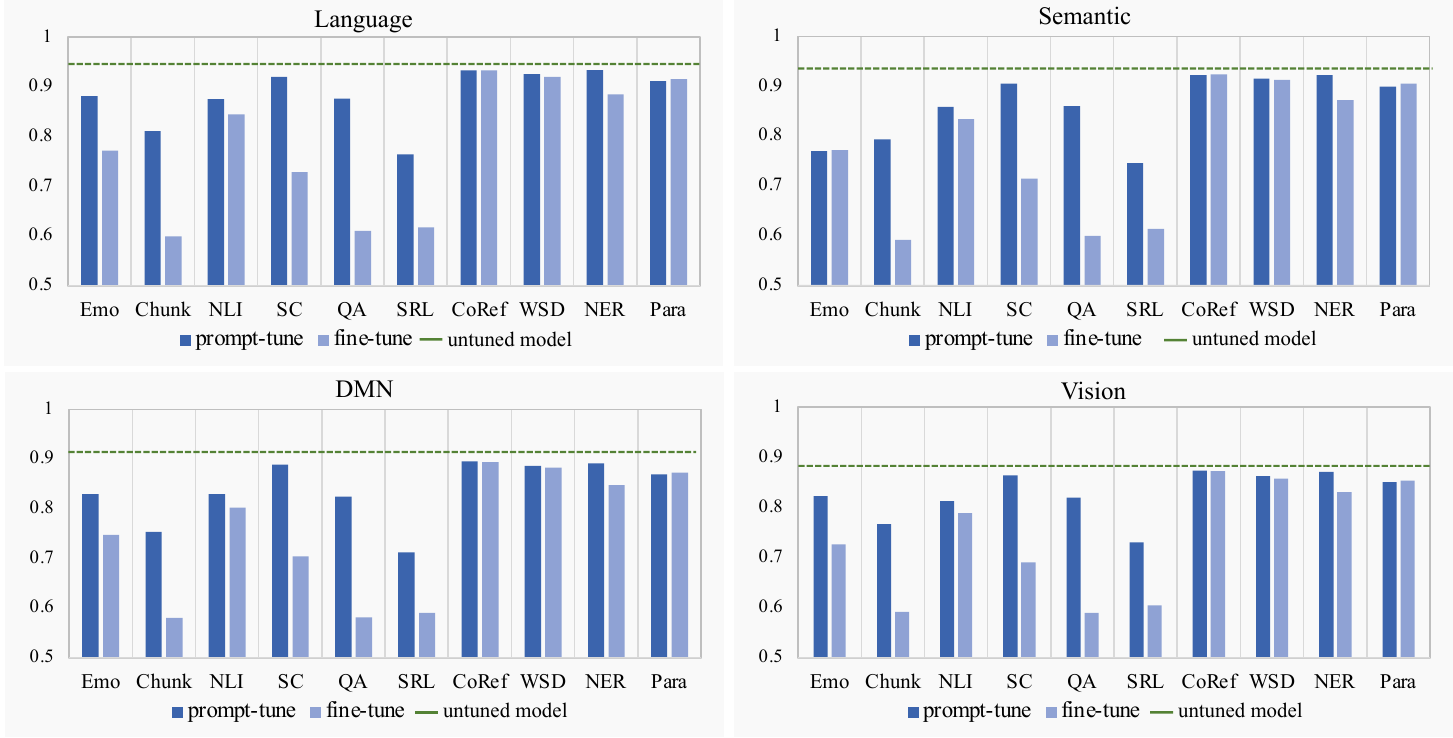}
\caption{Decoding accuracy evaluated by pairwise classification with task-supervised representations on the brain language networks, the ensemble semantic system, DMN and visual networks. The green dotted line denotes the decoding performance of the original untuned BERT representations.}
\label{fig3}
\end{figure*}

\begin{figure}[ht]
\centering
\includegraphics[width=3.35 in]{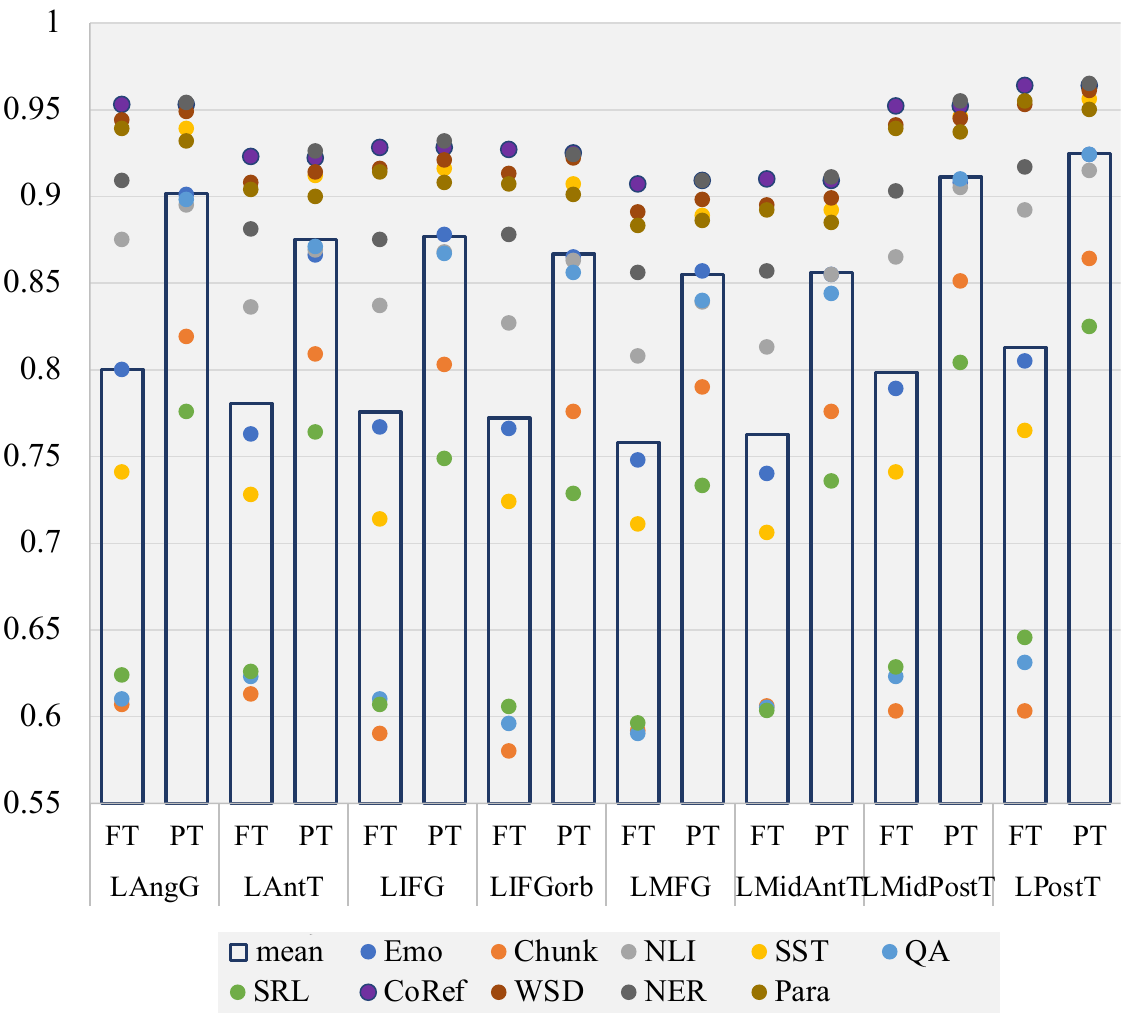}
\caption{ Decoding accuracy measured by pairwise matching with ROIs comprising the Language network. \textit{FT} and \textit{PT} respectively denote fine-tuning and prompt-tuning. \textit{mean} denotes the average score of all the decoding accuracies with fine-tuned or prompt-tuned representations. 
 }
\label{sheet1}
\end{figure}

 \begin{figure}[ht]
\centering
\includegraphics[width=2.9in]{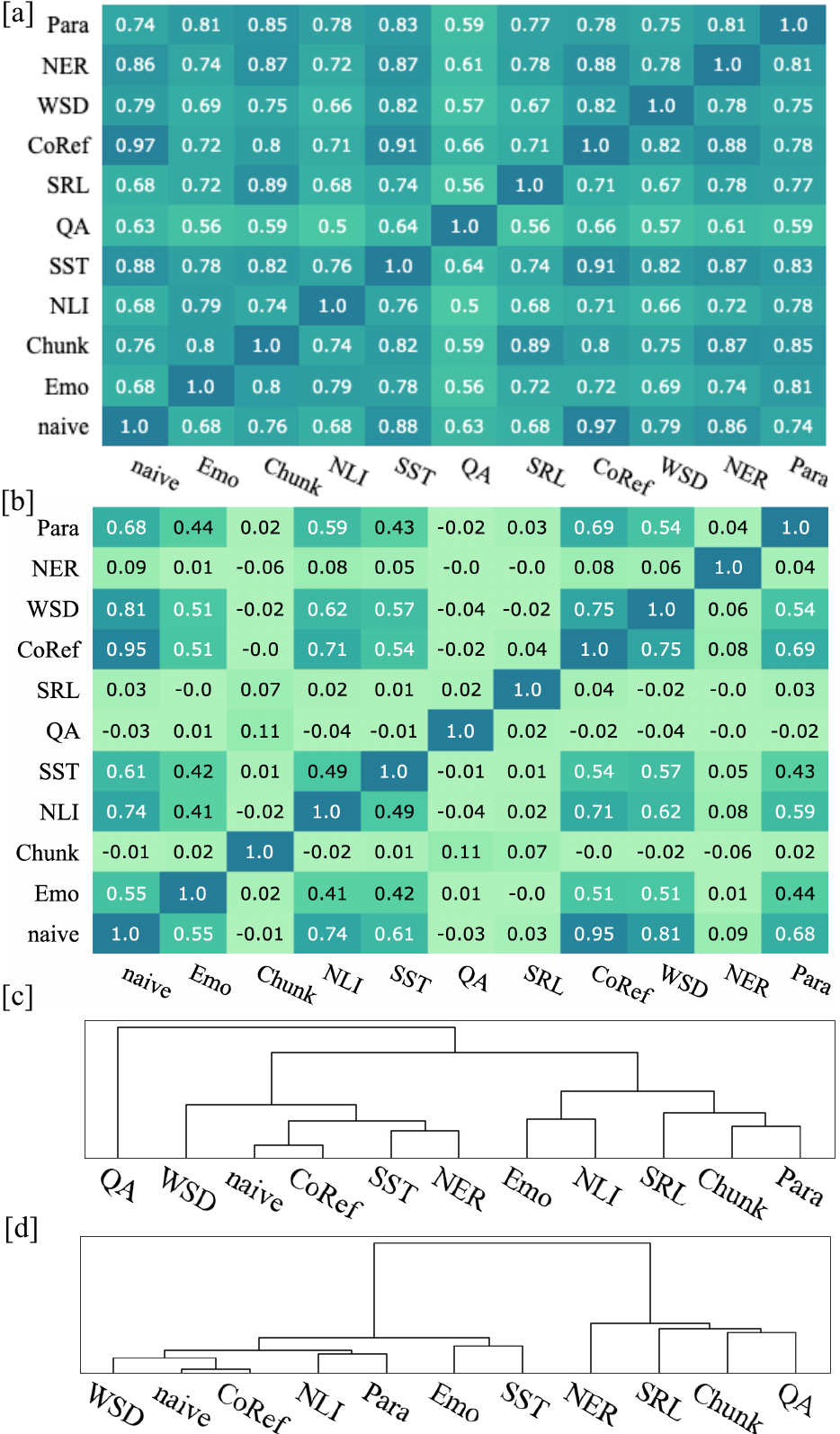}
\caption{[a] and [b]: Pearson correlation between each two of the supervised representations, [a] for the prompt-tuned representations and [b] for the fine-tuned representations.
[c] and [d]: Dendrograms constructed with hierarchical clustering on the Pearson correlations in [a] and [b], [c] for prompt-tuned representations and [d] for the fine-tuned representations.
 }
\label{line}
\end{figure}

The pre-trained Transformer model employed in this study is the open-source BERT-base-cased. 
As shown in Figure 2, task-specific tuning of the BERT model with fine-tuning and prompt-tuning are respectively conducted on each of the 10 tasks. 
These tuned models generate the supervised sentence embeddings needed for this study.
Sentences presented for collecting the brain images are input to the tuned models one by one. The hidden states of the last layer are averaged as sentence embeddings. The generated sentence embeddings have a dimension of 768.
We train on 4 Nvidia 3080TI GPUs with 12GB VRAM. 
Hyper-parameter searching is conducted to achieve optimal task performance. 
The influences of hyper-parameters settings including the size of prompts on the task performance are recorded in the appendix.

\section{Experimental Results}

 In this section, we compare the decoding performance of fine-tuned and prompt-tuned representations. We demonstrate how supervised representations trained on the 10 tasks differently decode the voxels of 4 major brain functional networks and the regions of interests (ROIs). We analyse the representational similarity structure of the supervised representations to explain the experimental results. We validate the findings from analyzing the decoding performance by paired t-test with the Bonferroni correction and report the results throughout with a significance level of $\alpha = 0.01$.

\subsection{Decoding Within Functional Networks}

Following previous work \cite{pereira2018toward,sun2020neural}, we select (i) the fronto-temporal language selective network \cite{fedorenko2011functional}, (ii) the ensemble semantic system  \cite{binder2009semantic}, (iii) the visual network 
 \cite{buckner2008brain} (including scene, face, body and object network) and (iii) the default mode network (DMN\footnote[4]{DMN is a network known for being active when the brain is at wakeful rest or during detailed thoughts related to external task performance.}). 
We conduct neural decoding on the voxels from each of the ROI comprising the functional networks and average the decoding performance across the entire network. 

As shown in Figure 3, we first observe a consistent and obvious advantage of prompt-tuning representations over fine-tuned peers. 
Specifically, we find a stark contrast in average decoding accuracy between fine-tuned and prompt-tuned representations in the tasks requiring reasoning with the syntactic structure (Chunk), and tasks analyzing the sentiment and emotion (SC and Emo). The accuracy differences are significant across the scanned subjects ($p<10^{-3}$). On decoding the language and semantic network, although 
representations fine-tuned on the WSD task slightly outperform the prompt-tuned one in average decoding accuracy, their difference does not stand up to the significance test among the scanned subjects ($p>10^{-2}$).

There are also tasks whose representations learned by fine-tuning and prompting yield similar decoding performance.
For both prompt-tuning and fine-tuning, representations learned from tasks dealing with concept meaning (CoRef, WSD and NER) deliver impressive decoding accuracy across the 4 functional networks. On decoding the Language and Semantic network, learning the CoRef and WSD task yields average decoding accuracies higher than other tasks except NER across subjects for both types of tuned representations. The difference between WSD and CoRef is not consistently significant ($p>10^{-2}$). 
On decoding the visual network and DMN, learning the CoRef task yields the best average decoding accuracy.

\subsection{Decoding Within Regions of Interests}

In this subsection, we present the decoding performance of fine-tuned and prompt-tuned representations on the ROIs comprising the language network.\footnote[5]{There are 8 ROIs comprising the language network and their full names are detailed in the appendix.} We conduct neural decoding on each ROI for all the subjects and show the averaged decoding performance among the subjects in Figure 4. 

Though fine-tuning on CoRef and WSD yields representations that better decode some ROIs than prompt-tuning, after pooling across all the tasks, we find that on all the ROIs prompt-tuning exceeds fine-tuning in average decoding accuracy.  
For both fine-tuned and prompt-tuned representations, left posterior temporal cortex (LPostTemp) is the ROI where the highest average decoding accuracy is observed while left middle frontal gyrus (LMFG) is the lowest. 
Actually, if we rank different ROIs according to the average decoding accuracy, fine-tuning and prompt-tuning produce the same rank
which is also consistent with the decoding performance of the naive BERT representations. 
We also find that on different ROIs, the supervised representations show different decoding performance patterns. For example for the fine-tuned representations, the CoRef task representations yield the best decoding accuracy on most language network ROIs. But for the prompt-tuned embeddings, the WSD task representations outperform.     

\subsection{Representational Similarity of Supervised Embeddings}

In this subsection, we dive deep into the similarity relationship of supervised representations.
We calculate the Pearson's correlation between each two of the task-supervised representations and depict the correlation as heatmaps in 
Figure 5[a] (prompt-tuned representations) and Figure 5[b] (fine-tuned representations).

As shown in Figure 5, the fine-tuned representations largely diverge in similarity with the naive untuned BERT representations, while prompt-tuned representations correlate closer to the untuned representations.  Representations fine-tuned on the QA, Chunk and SRL task even show zero or negative correlation with the untuned representations, in line with previous work \cite{kumar2022finetuning} which proved fine-tuning can distort pre-trained features. 
Neural decoding accuracies with fine-tuning on the QA, Chunk and SRL tasks are also the lowest three observed in previous experiments, as depicted in Figure 3 and 4.
Both fine-tuning and prompt-tuning on the CoRef task yield supervised representations that accurately decode brain activities.  The CoRef representations with fine-tuning and prompt-tuning also show top correlations with the untuned model.

To gain a deeper insight into the relationship among the supervised representations, we conduct a hierarchical clustering and illustrate the results as dendrograms in Figure 5[c] and 5[d]. For both fine-tuning and prompt-tuning, CoRef representations share one cluster with the naive untuned representations. Chunk and SRL representations are closely clustered, distant from the naive and CoRef representations in the hierarchy. 
We further calculate the variance of neural decoding accuracies explained by the correlation between each of the supervised representations and the naive BERT representation. 
We find that such correlation explains 35.35\% variance of prompt-tuned representations' neural decoding accuracy within the brain language network, and explains 44.22\% variance of neural decoding accuracy within the ensemble semantic system.

\section{Discussion}

In this section we summarize the experimental findings and discuss possible explanations for the findings.
First, we find that decoding performance with fully fine-tuned representations do not significantly exceed prompt-tuned representations on decoding any of the 4 functional network voxels. 
Full fine-tuning updates the entire set of model parameters to fit on a downstream NLU task, which has been proven to distort the pre-trained features.
However, the human brain does not need to reformulate an entire functional network to learn a single new task \cite{cole2013multi,fedorenko2011functional}. Learning from a new task also does not necessarily lead to distortion of previously acquired knowledge \cite{parisi2019continual}. 
In this sense, full fine-tuning is cognitively inconsistent with the human brain's mechanism of language representation.
Prompt-tuning, in contrast with fine-tuning, freezes the pre-trained weights and learns only trainable embeddings to guide the base model to fit on a downstream task. 
As hypothesized, prompt-tuned embeddings better decode the information represented in the human brain for language understanding than fine-tuned peers. 

Second, we find that for both fine-tuning and prompt-tuning, representations tuned on the CoRef and WSD task yield higher average decoding accuracy than other tasks, while the Chunk task yields the lowest.
The CoRef and WSD tasks require complex reasoning involving concept meaning and semantic relation, while the Chunk task requires modeling the shallow syntactic structure.
This indicates that deep level semantic information could be involved in human brain's language representation, possibly taking a larger proportion than the shallow syntactic information.
We also find that the neural decoding performance of the prompt-tuned representations can be at least partially explained by their correlation with the untuned naive model.
This means that general domain knowledge learned by the pre-trained model plays an important role in decoding the brain representations of language.

Third, we observe that representations tuned on half of the evaluated tasks do not significantly improve over the untuned model in neural decoding accuracy.  
Learning task-specific features leads to updating or even reformation of the general domain knowledge acquired in pre-training, which could be important for decoding human language representations.
If the task-specific feature learned from tuning is not as important as the general domain knowledge in decoding brain representations, we can expect tuning yields decreased decoding accuracy.
Since the tasks are from popular NLP benchmarks and the models are tuned with the hyper-parameter search to achieve their optimal task performance, this finding also suggests these benchmarks might not robustly and accurately reflect how well the tuned models resemble the brain's mechanism for language understanding. 
It then urges further studies to build benchmarks that assess the cognitive and biological plausibility of the task-supervised models.

\section{Conclusion}

In this study, for the first time we propose to link the human brain and task-supervised artificial representations by prompt-tuning. We take 10 NLU tasks involving diverse linguistic-cognitive skills from  widely used NLU benchmarks. A pre-trained BERT model is fitted on the 10 tasks with full fine-tuning and prompt-tuning, respectively. The supervised representations generated by the tuned models are then employed to decipher the brain activation patterns through neural decoding. We comprehensively compare the prompt-tuned supervised representations against the fine-tuned peers on 4 functional networks and their comprising ROIs. The comparison demonstrates that prompt-tuned representations could better decode the neural responses to a linguistic stimulus than fully fine-tuned representations in most cases. We also demonstrate how task representations differently decode brain activities.
We believe that the proposed prompt-tuning-based framework and experimental findings will help to better understand the relationship between supervised artificial and brain language representations.

\section*{Acknowledgements}
This work is funded by the CALCULUS project (European Research Council Advanced Grant H2020-ERC-2017-ADG 788506). We thank Dr. Shaonan Wang very much
for proof-reading.

\bibliographystyle{named}
\bibliography{ijcai23}

\end{document}